\documentclass[a4paper,fleqn]{cas-dc}

\usepackage[authoryear,longnamesfirst]{natbib}
\usepackage{subfigure}
\usepackage{caption}
\usepackage{diagbox}
\usepackage{amsmath}
\usepackage{amssymb}
\usepackage{url}
\usepackage{hyperref}
\usepackage{color}
\usepackage{xcolor}
\usepackage{textcomp}
\usepackage{stfloats}
\usepackage{url}
\usepackage{verbatim}
\usepackage{graphicx}
\usepackage{cite}
\usepackage{multirow}
\usepackage{makecell}
\usepackage[ruled]{algorithm2e}
\usepackage{algpseudocode}

\newcommand{\modified}[1]{{\color{black} #1}}

\def\tsc#1{\csdef{#1}{\textsc{\lowercase{#1}}\xspace}}
\tsc{WGM}
\tsc{QE}
\tsc{EP}
\tsc{PMS}
\tsc{BEC}
\tsc{DE}

\begin{document}
\let\WriteBookmarks\relax
\def\floatpagepagefraction{1}
\def\textpagefraction{.001}

\shorttitle{FedMed-GAN}

\shortauthors{Jinbao Wang et~al.}

\title [mode = title]{FedMed-GAN: Federated Domain Translation on Unsupervised Cross-Modality Brain Image Synthesis}                      

\author[1]{Jinbao Wang}\fnmark[1]\ead{linkingring@163.com}
\credit{Conceptualization, Methodology, Data Curation, Validation, Visualization, Writing - original draft \& editing}
\author[1,2]{Guoyang~Xie}\fnmark[1]\ead{guoyang.xie@surrey.ac.uk}
\credit{Conceptualization, Methodology, Data Curation, Investigation, Writing - original draft \& editing}
\author[3]{Yawen Huang}\fnmark[1]\ead{yawenhuang@tencent.com}
\credit{Conceptualization, Methodology, Investigation, Writing - review \& editing, Supervision}
\author[4]{Jiayi Lyu}\ead{lyujiayi21@mails.ucas.ac.cn}
\credit{Data Curation, Visualization, Writing - review \& editing}
\author[1,5]{Feng Zheng}\cormark[1]\ead{zhengf@sustech.edu.cn}
\credit{Supervision, Project administration, Funding acquisition, Writing - review \& editing}
\author[3]{Yefeng Zheng}\ead{yefengzheng@tencent.com}
\credit{Supervision, Writing - review \& editing}
\author[2,6]{Yaochu~Jin} \cormark[1]\ead{yaochu.jin@uni-bielefeld.de}
\credit{Supervision, Writing - review \& editing}

\cortext[cor1]{Corresponding authors.}
\fntext[fn1]{These authors contributed equally to this work.}

\address[1]{Department of Computer Science and Engineering, Southern University of Science and Technology, 518055, China}
\address[2]{Department of Computer Science, University of Surrey, Guildford, Surrey GU2 7XH, U.K.}
\address[3]{Tencent Jarvis Lab, Shenzhen 518040, China}
\address[4]{School of Engineering Science, University of Chinese Academy of Sciences, Beijing, China}
\address[5]{Peng Cheng Laboratory, Shenzhen 518000, China}
\address[6]{Faculty of Technology, Bielefeld University, 33619 Bielefeld, Germany}

\begin{abstract}
Utilizing multi-modal neuroimaging data is proven to be effective in investigating human cognitive activities and certain pathologies. However, it is not practical to obtain the full set of paired neuroimaging data centrally since the collection faces several constraints, \textit{e.g.}, high examination cost, long acquisition time, and image corruption. In addition, these data are dispersed into different medical institutions and thus cannot be aggregated for centralized training considering the privacy issues. There is a clear need to launch federated learning and facilitate the integration of dispersed data from different institutions. In this paper, we propose a new benchmark for federated domain translation on unsupervised brain image synthesis (FedMed-GAN) to bridge the gap between federated learning and medical GAN. FedMed-GAN mitigates the mode collapse without sacrificing the performance of generators, and is widely applied to different proportions of unpaired and paired data with variation adaptation properties. We treat the gradient penalties using the federated averaging algorithm and then leverage the differential privacy gradient descent to regularize the training dynamics. A comprehensive evaluation is provided for comparing FedMed-GAN and other centralized methods, demonstrating that the proposed algorithm outperforms the state-of-the-art. \modified{Our code is available at: \hyperref[https://github.com/M-3LAB/FedMed-GAN]{https://github.com/M-3LAB/FedMed-GAN}.}

\end{abstract}



\begin{keywords}
Federated Learning \sep Unsupervised Learning \sep Cross-Modality Synthesis \sep  Brain Image \sep Deep Learning

\end{keywords}

\maketitle

\section{Introduction}
The majority of existing medical datasets \citep{Siegel2019CancerS2, Bakas2017BrainlesionGM}, especially for neuroimaging data, are high-dimensional and heterogeneous. \modified{For instance, positron emission tomography (PET) and magnetic resonance imaging (MRI) are imaging techniques that measure information about organs and tissues to aid in diagnosis or treatment monitoring.} These pair of multi-modal data provide more complementary information to investigate certain pathologies and neurodegeneration. However, it is infeasible to acquire a full set of paired multi-modal neuroimaging data. There are two issues. \modified{First, collecting multi-modal neuroimaging data is very expensive; for instance, a normal MRI in New York can cost more than one thousand dollars and more than eight thousand RMB in Beijing. In addition, incorrect patient motion results in neuroimage corruption. Retaking neuroimaging for each individual patient costs approximately one hour. Therefore, it is challenging to obtain a complete set of paired neuroimages. Second, sharing neuroimages across multi-hospitals needs to be authorized by each patient. Furthermore, many medical institutions are not allowed to share their data due to local hospital regulations, even if the identifiable information has been removed for protecting the privacy of patients. Even in multi-hospital collaborative research, they would undertake integrated analysis rather than sharing research data with other hospitals. Hence, data isolation and privacy concerns are the fundamental problems hampering a large-scale and multi-institute neuroimaging research.}

\textbf{Cross-Modality Brain GAN.}
Generative adversarial networks (GANs) \citep{goodfellow2014generative} are state-of-the-art deep generative models, which have achieved huge success in image synthesis. GANs alternatively train two networks, in which the generator maps a random input vector into a high-dimensional space and the discriminator judges the outputted data from real ones. These two networks aim to defeat each other. The training objective of the generator is to create a ``fake'' content to confuse the discriminator, while the objective of the discriminator is to improve distinguishable ability. GANs need a large amount of paired data for training, but most of the paired neuroimaging data are scattered in different hospitals. Due to privacy legislation, it is not feasible to aggregate the full set of paired neuroimages from various medical institutions. \modified{State-of-the-art cross-modality neuroimage synthesis methods~\citep{Yang2021AUH, Huang2020MCMTGANMC} do not target on the neuroimage data distributed into different medical institutions. Hence, we propose FedMed-GAN to address these fundamental problems by simulating as much as possible proportions of unpaired and paired data of each client with a variety of data distributions for all clients. Moreover, we construct a comprehensive benchmark and thoroughly investigate the issues of mode collapse and performance drop of FedMed-GAN, facilitating the development of federated cross-modality neuroimage synthesis.}  

\textbf{Federated GAN.}
Recently, a large amount of effort has been made to facilitate the availability of medical data without violating the privacy issue. Federated Learning (FL) is one of the popular approaches.       
FL is a decentralized approach where local clients train their local models without transmitting data to a central server, and the global model aggregates the gradients from clients \citep{mcmahan2017communication}. In addition, FL with GANs has witnessed some pilot progress on image synthesis \citep{chen2020gs}.
For example, DP-FedAvg-GAN~\citep{DBLP:conf/iclr/AugensteinMRRKC20} trains GANs with the differential privacy-preserving algorithm, which clips the gradients to bound sensitivity and adds calibrated random noise to introduce stochasticity.
\modified{Specifically, DP-FedAvg-GAN locates a discriminator for each client and a generator on the server. Each client updates its discriminator by using its own dataset and creating a fake image for the global generator.} In this case, the generator is only exposed to the discriminator and never uses the real data. However, each client can only assess one domain data, which cannot fully leverage the existing training paradigm of GANs, especially for CycleGAN \citep{zhu2017unpaired}. 
In addition, we also discover that DP-FedAvg-GAN may suffer from the generator mode collapse and slow down the convergence rate in our FedMed-GAN benchmark. The performance of DP-FedAvg-GAN is lower than that of the centralized training due to the differential privacy guarantee. \modified{Our method utilizes the shared features across multi-modal neuroimages to guide the generator, which could largely mitigate the side-effects of DP-FedAvg-GAN.}

Our contributions can be summarized as follows:
\begin{itemize}
\item \modified{FedMed-GAN, to the best of our knowledge, is the first work to establish a new benchmark for federally cross-modality brain image synthesis, which greatly facilitates the development of medical GAN with differential privacy guarantees.}
\item \modified{We provide comprehensive explanations for addressing mode collapse and performance drop compared to centralized training.}
\item The proposed work simulates as much as possible proportions of unpaired and paired data for each client with various data distributions for all clients. The performance of FedMed-GAN remains stable when facing long-tail data distributions.
\end{itemize}

\section{Related Work}
\textbf{Cross-Modality Medical Image Synthesis.}
\modified{Existing medical image-to-image translation~\citep{jiang2019synthesize,ren2021segmentation,kong2021breaking} has demonstrated their considerable research and clinical analysis potential.} Of these methods, supervised GANs are still the mainstream for cross-modality neuroimaging data synthesis~\citep{Wang2018LocalityAM,Dar2019ImageSI,Sharma2020MissingMP,Yu2020SampleAdaptiveGL,Yu20183DCB,Zuo2021DMCFusionDM}. However, synthesizing in a supervised manner requires paired data for training, which is difficult to implement in practice. To solve this problem, both semi-supervised and unsupervised methods are then launched to eliminate the need of paired data. \modified{\citep{Guo2021AnatomicAM} leverage a lesion segmentation network as a teacher to guide the generator by using unpaired training data. \citep{Shen2021MultiDomainIC} and \citep{Zhou2021AnatomyConstrainedCL} also utilize the high-level tasks to guide the cross-modality image synthesis. Huang~\textit{et al.}~\citep{Huang2020SuperResolutionAI, Huang2020MCMTGANMC} make full use of unpaired cross-modality data and project them into a common space. The attributed features from the common space bring great helpful to synthesize the missing target modality data. \citep{Li2022AGF} employ the dual-domain attention mechanism to extract highly discriminative features on the lesion area. \citep{Wu2023AGGNAG} propose a MRI oriented novel attention-based glioma grading network into mutli-scale feature extraction process, which aims to promote the synergistic interaction among different modality information.} \citet{kong2021breaking} introduce a new I2IT model called RegGAN, which converts the unsupervised I2IT task into a supervised I2IT with noisy labels. However, RegGAN cannot deal with the severe distortion, which probably happens in the realistic scenarios.

\textbf{Federated GAN.}
\modified{Data isolation and privacy concerns are the fundamental challenges to multi-institute, large-scale neuroimaging research. Federated learning, as a privacy-preserving decentralization strategy, enables clients to train their own models without transmitting data to a central server by aggregating client progresses to update a global model~\citep{DBLP:conf/aistats/McMahanMRHA17,DBLP:conf/icml/YurochkinAGGHK19,DBLP:conf/mlsys/LiSZSTS20,DBLP:conf/iclr/WangYSPK20}.} However, directly incorporating the generative adversarial framework into the federated learning is challenging, because the cost functions may not converge using federated gradient aggregation in a min-max setting between the discriminator and the generator~\citep{DBLP:conf/iclr/AugensteinMRRKC20,DBLP:conf/nips/ChenOF20,DBLP:journals/corr/abs-2106-09246}. \modified{DP-SGD-GAN~\citep{DBLP:conf/iclr/AugensteinMRRKC20} provides a differential privacy-preserving algorithm, which clips the gradients to bound sensitivity and adds calibrated random noise to introduce stochasticity. DP-SGD heavily relies on the fine-tuning of the clipping bound of the gradient norm. Specifically, the optimal clipping bound is sensitive and varies greatly with the model architecture, making the implementation of DP-SGD difficult. G-PATE~\citep{Long2019GPATESD} is similar to our work, but it only trains the generator with DP guarantee. Thus, G-PATE framework incurs high privacy cost. GS-WGAN \citep{DBLP:conf/nips/ChenOF20} enables the release of a sanitized version of sensitive data while maintaining stringent privacy protections. This method can distort gradient information, allowing a deeper model to be trained with more informative samples. In contrast to centralized training, they do not investigate the mode collapse and performance drop issues in depth. Despite that these methods demonstrate excellent performance on a variety of tasks, cross-modality image synthesis remains unexplored, and a theoretical or empirical analysis of convergence still lacks.}

\begin{figure*}[htbp]
	\centering
    \includegraphics[width=0.95\textwidth]{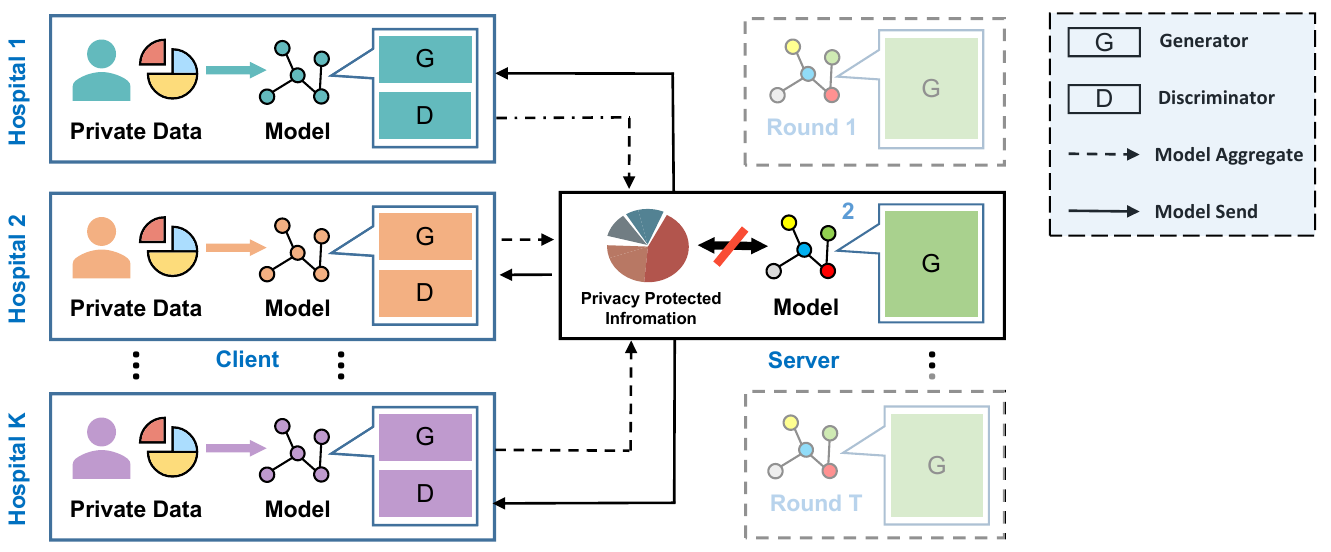}
	\caption{
		The pipeline of FedMed-GAN. The generator of each client participates in federated learning, while the discriminator of each client does not.
	}\label{fig:pipeline}
\end{figure*}

\section{FedMed-GAN}

\textbf{Federated Model Setup.}\label{fed_model_setup}
\modified{The federated setting of our generator and discriminator is described in Fig.~\ref{fig:pipeline}. We employ CycleGAN~\citep{zhu2017unpaired}, MUNIT~\citep{huang2018multimodal}, UNIT~\citep{DBLP:conf/nips/LiuBK17}, HLHGAN~\citep{Huang2020SuperResolutionAI}, and Hyper-GAN~\citep{Yang2021AUH}  as our baseline models. Our baseline methods have two generators $G_{A \leftrightarrow B}$, discriminators $D_{A \leftrightarrow B}$. $G_{A \rightarrow B}$ generates B-modal images from A-modal samples. $D_{A \rightarrow B}$ distinguishes whether the generated B-modal data from A-modal samples is fake. Two generators ($G_{A \rightarrow B}$, $G_{B \rightarrow A}$) of each client joins in the aggregation process of FedAvg~\ref{alg:server_g}. In other words, the generators are separately aggregated into the server's generators. Then, the server sends its generators into different hospitals. Moreover, according to GS-WGAN~\citep{chen2020gs}, locating the discriminators ($D_{A \rightarrow B}$, $D_{B \rightarrow A}$) of each client locally can further increase the level of privacy preserving. Thus, the discriminators ($D_{A \rightarrow B}$, $D_{B \rightarrow A}$) are retained locally and do not participate in aggregation process.}

\begin{figure}[htbp]
    \centering
    \includegraphics[width=0.5\textwidth]{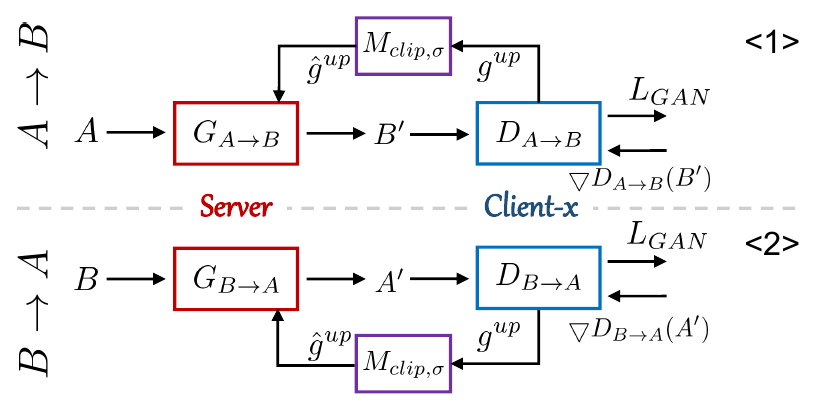}
	\caption{The architecture of FedMed-GAN, equipped with CycleGAN by adopting DP. The generator $G$ indicates that the models are shared with servers and clients. The discriminator $D$ indicates that the models are located in the client without sharing with one another. $M$ denotes the DP-SGD operator. $g^{up}$ denotes the back-propagation gradient from discriminators to generators. $\hat{g}^{up}$ denotes the clipped gradient after $M$.  
	}\label{fig:fed_dp}
\end{figure}

\textbf{Differential Privacy.}
Considering the superiority of DP-SGD \citep{abadi2016deep} and GS-WGAN \citep{DBLP:conf/nips/ChenOF20}, we provide a differential privacy (DP) mechanism and show the detailed architecture in Fig.~\ref{fig:fed_dp}. The definition of differential privacy~\citep{Dwork2014TheAF} is defined as below:

\textit{Definition 1}. A random mechanism $\mathcal{A}_{p}$ satisfies $(\epsilon, \delta)$-differential privacy if for any output's subset ($\mathcal{S}$) and any two datasets $\mathcal{M}$, $\mathcal{M^{'}}$, the following probability inequality holds: 
\begin{equation}\label{eq:privacy_budget}
    \mathbb{P}(\mathcal{A}_{p}(\mathcal{M} \in \mathcal{S}) \leq e^{\epsilon} \cdot \mathcal{P}(\mathcal{A}_{p} (\mathcal{M^{'}} \in \mathcal{S}) + \delta,
\end{equation}
where $\delta>0$ and $\epsilon$ is the privacy budget indicating the privacy level, i.e., a smaller value of $\epsilon$ implies stronger privacy protection.
To prevent privacy leakage from the server's generators, we train the client's generators ($G_{A \rightarrow B}$, $G_{B \rightarrow A}$) in a privacy-preserving manner. DP-SGD \citep{abadi2016deep} is adopted by using our DP. 
To realize DP, DP-SGD clips the gradient to bound sensitivity and add a calibrated random noise to induce stochasticity. 

We directly apply DP-SGD to the parameters $\theta_{G}$ of the generators since the discriminator does not participate in the federated learning stage,
\begin{equation}
    \begin{aligned}
          g_{G}^{t} &= \nabla_{\theta_{G}} \mathcal{L_{G}}(\theta_{G}^{t}, \theta_{D}^{t}), \\
          \hat{g}_{G}^{t} &= M_{clip, \sigma} = clip(g^{t}, C) + \mathcal{N}(0, \sigma^{2}C^{2}I), \\
          \theta_{G}^{t+1} &= \theta_{G}^{t} - \eta_{G} \hat{g}_{G}^{t},
    \end{aligned}
\end{equation}
where $t$ is the epoch number in training. $C$ denotes the clip bound of the gradient. $\sigma$ represents the standard deviation of Gaussian noise. $g_{G}$ is the back-propagation gradient from the generator loss. $\hat{g}_{G}$ is the clipped gradient after $M_{clip, \sigma}$.
The relationship between the noise variance and differential variance is given as follows:
\begin{equation}\label{eq:privacy_budget}
    \sigma = \frac{2q}{\epsilon}\sqrt{n_{d}\textrm{log}\frac{1}{\delta}},
\end{equation}
where $q$ and $n_{d}$ denote the sample probability for each instance and the total batch number of the local dataset, respectively.

\textbf{Loss Function}
For the synthesis process: $A \rightarrow B$ and its discriminator $D_{B}$, its generator loss function is defined as:
\begin{equation}\label{eq:g_loss}
\mathcal{L}_{G} = \mathbb{E}_{a \sim p_{data}(a)} \left [ \textrm{log}(1-D_{B}(G(a))) \right ].
\end{equation}
And the discriminator loss function is defined as:
\begin{equation}\label{eq:d_loss}
    \mathcal{L}_{D} = \mathbb{E}_{b \sim p_{data}(b)}  \left [ \textrm{log} D_{B}(b) \right ].
\end{equation}
For the synthesis process: $B \rightarrow A$, its generator loss function and discriminator loss are similar to Eq.~(\ref{eq:g_loss}) and Eq.~(\ref{eq:d_loss}).
The cycle-consistency loss function from~\citep{zhu2017unpaired} is defined as below: 
\begin{equation}\label{eq:cycle}
\begin{aligned}
      \mathcal{L}_{cycle}(A, \hat{A}, B, \hat{B}) = \mathbb{E}_{a \sim p_{data}(a)}  \left [ \left\| G_{2}(G_{1}(a)) - a \right\| \right ]_{1}\\
    + \mathbb{E}_{b \sim p_{data}(b)}  \left [ \left\| G_{1}(G_{2}(b)) - b \right\| \right ]_{1}  .
\end{aligned}
\end{equation}

\section{Misaligned Unpaired Data (MUD)}
\begin{figure}[htb]
    \centering
    \includegraphics[width=0.45\textwidth]{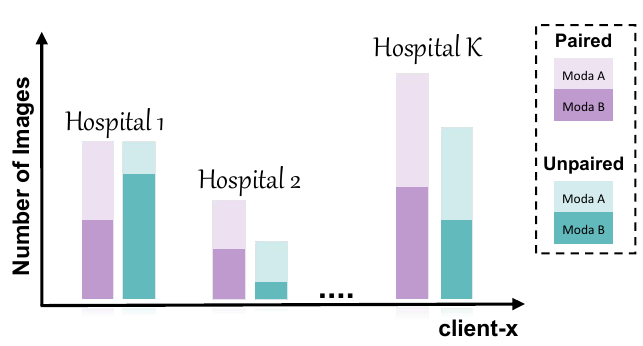}
	\caption{Examples of data distribution and performance during training.}\label{fig:data-spitted}
\end{figure}

We address the task of federated self-supervised learning for MUD by formulating several realistic settings. To simulate the real data distribution as much as possible, we adopt the following settings as shown in Fig.~\ref{fig:data-spitted}. 

Firstly, we divide ready-made data into several clients, where each client has its own private and unique data. Meanwhile, these clients (hospitals) contain unbalanced multi-modal images, i.e. having different numbers of patients in each client or having different numbers of images in each modality. Secondly, without losing generality, we explore both paired and unpaired cases in our experiments. For example, we randomly select the multi-modal slices from the same patient as paired data like in Hospital 1 or select them from patients with different IDs as unpaired data like in Hospital 2. After that, we randomly transform the input images at a certain threshold of rotation, translation and scaling, and there is the majority of misaligned images in the constructed training set. 

\section{Algorithm}
\modified{The aggregation algorithm for the server's generator is denoted in Algorithm~\ref{alg:server_g}, the client's generator optimization algorithm is described in Algorithm~\ref{alg:client_g}, and the client's discriminator algorithm is described in Algorithm~\ref{alg:client_d}.}

\begin{algorithm}[th]
\small
\caption{Server-orchestrated training loop}\label{alg:server-training}
\KwData{total number of hospitals $N\in \mathbb{N}$, total number of rounds of $T \in \mathbb{N}$, $G(A \rightarrow B)$: $G1$, $G(B \rightarrow A)$: $G2$, $D(A \rightarrow B)$: $D_{1}$, $D(B \rightarrow A)$: $D_{2}$: clients $\theta_{n}$, server $\theta_{S}$}

\textbf{Initialize:} gen. $\theta^{0}_{G1}$ and $\theta^{0}_{G2}$, dis. $\theta^{0}_{D1}$ and $\theta^{0}_{D2}$

\For{each round $t$ from $0$ to $T$}{
    \For{each hospital $n \in N$ in parallel}{        
        $\theta^{t+1}_{n:G1}, \theta^{t+1}_{n:G2} \leftarrow \mathrm{ClientGenUpdate}(\theta^{t}_{G1}, \theta^{t}_{D1},\theta^{t}_{G2}, \theta^{t}_{D2})$ \\
        $\theta^{t+1}_{n:D1}, \theta^{t+1}_{n:D2} \leftarrow \mathrm{ClientDiscUpdate}(\theta^{t}_{D1}, \theta^{t}_{D2})$ \\
        }
        $\theta^{t+1}_{S:G1}, \theta^{t+1}_{S:G1} \leftarrow \mathrm{Aggregate}(\theta^{t+1}_{n:G1}, \theta^{t+1}_{n:G2})$ \\
}
\textbf{Return:} server generators $\theta_{G1}$, $\theta_{G2}$
\label{alg:server_g}
\end{algorithm}

\begin{algorithm}[th]
\small
\caption{ClientGenUpdate($\theta_{D1}$, $\theta_{G1}$, $\theta_{D2}$, $\theta_{G2}$)}\label{alg:genupdate}
\KwData{total number of steps $ J \in \mathbb{N}$, batch size $S \in \mathbb{R}^+$, learning rate $\eta_G \in \mathbb{R}^+$, gen. input size $n_U \in \mathbb{N} $, $A_{fake}=A'$, $B_{fake}=B'$,  $G(A \rightarrow B)$: $G_{1}$, $G(B \rightarrow A)$: $G_{2}$, $\hat{A} = G_{2}(B')$, $\hat{B} = G_{1}(A')$, clip bound $C$, noise standard deviation $\sigma$, gradient sensitization mechanism $\mathcal{M}$, }

\textbf{Loss function:}$L_{G_{1}}(\theta_{G1}, \theta_{D1}; B')$, $L_{G_{2}}(\theta_{G2}, \theta_{D2}; A')$, $L_{cycle}(A, \hat{A}, B, \hat{B})$
                        
\textbf{Initialize:} $\theta_{G1} \leftarrow \theta^0_{G1}$, $\theta_{G2} \leftarrow \theta^0_{G2}$, $s \leftarrow $ $k$'s paired or unpaired data split into $S$ batches

\For{each generator training step $j$ from $1$ to $J$}{
    $U_{A} \leftarrow $ (sample $S$ random vectors of dim. $n_U$) \\
    $U_{B} \leftarrow $ (sample $S$ random vectors of dim. $n_U$) \\
    $A' \leftarrow G(U_{B}, \theta_G)$ \\
    $B' \leftarrow G(U_{A}, \theta_G)$ \\
    $g_{1} = \bigtriangledown (\mathcal{L}_{G1}(\theta_{G1}, B'; \theta_{D1}) +  \mathcal{L}_{cycle})$ \\
    $g_{2} = \bigtriangledown (\mathcal{L}_{G2}(\theta_{G2}, A'; \theta_{D2}) +  \mathcal{L}_{cycle})$ \\
    $\hat{g}_{1} = \mathcal{M}(g_{1}) = \mathrm{clip}(g_{1}, C) + N(\sigma^2C^2I)$ \\
    $\hat{g}_{2} = \mathcal{M}(g_{2}) = \mathrm{clip}(g_{2}, C) + N(\sigma^2C^2I)$ \\
    $\theta_{G1} = \theta_{G1} - \eta_G \hat{g}_{1}$ \\
    $\theta_{G2} = \theta_{G2} - \eta_G \hat{g}_{2}$ \\
    }
\textbf{Return:} client generators $\theta_{G1}$, $\theta_{G2}$
\label{alg:client_g}
\end{algorithm}

\begin{algorithm}[t]
\small
\caption{ClientDiscUpdate($\theta_{D1}$, $\theta_{D2}$)}\label{alg:disupdate}
\KwData{number of steps $ J \in \mathbb{N}$, batch size $S \in \mathbb{R}^+$, learning rate $\eta_G \in \mathbb{R}^+$, gen. input size $n_U \in \mathbb{N} $, $A_{fake}=A'$, $B_{fake}=B'$,  $G(A \rightarrow B)$: $G_{1}$, $G(B \rightarrow A)$: $G_{2}$, $D(A \rightarrow B)$: $D_{1}$, $D(B \rightarrow A)$: $D_{2}$}

\textbf{Loss function: } $L_{D_{1}}( \theta_{D1};B,B')$, $L_{D_{2}}( \theta_{D2};A,A')$

\textbf{Initialize:} $\theta_{D1} \leftarrow \theta^0_{D1}$, $\theta_{D2} \leftarrow \theta^0_{D2}$, $s \leftarrow $ $k$'s unpaired data split into $S$ batches

\For{each generator training step $j$ from $1$ to $J$}{
    $U_{A} \leftarrow $ (sample $S$ random vectors of dim. $n_U$) \\
    $U_{B} \leftarrow $ (sample $S$ random vectors of dim. $n_U$) \\
    $B' \leftarrow G_{1}(U_{A}, \theta_{G1})$ \\
    $A' \leftarrow G_{2}(U_{B}, \theta_{G2})$ \\
    $\theta_{D1} = \theta_{D1} - \eta_D \bigtriangledown (\mathcal{L}_{D1}(\theta_{D1}; B, B')$\\
    $\theta_{D2} = \theta_{D2} - \eta_D \bigtriangledown (\mathcal{L}_{D2}(\theta_{D2}; A, A')$ \\
    }
\textbf{Return:} client generator $\theta_{D1}$ and $\theta_{D2}$
\label{alg:client_d}
\end{algorithm}

\section{Experiments}\label{sec:experiments}

\subsection{Implementations}

\textbf{IXI}\footnote{\url{https://brain-development.org/ixi-dataset}}~\citep{Aljabar2011ACM} collects nearly 600 MR images from normal and healthy subjects at three hospitals. The MR image acquisition protocol for each subject includes T1, T2, PD-weighted images (PD), MRA images, and Diffusion-weighted images (15 directions). Here, we only use T1 (581 cases), T2 (578 cases), and PD (578 cases) data to conduct our experiments, and select the paired data with the same ID from the three modes. The image has a non-uniform length on the z-axis with the size of $ 256 \time 256 $ on the x-axis and the y-axis. The IXI data set is not divided into a training set and a test set. Therefore, we randomly split the whole data as the training set (0.8) and the test set (0.2).

\textbf{BraTS2021}\footnote{\url{http://www.braintumorsegmentation.org}}~\citep{Siegel2019CancerS2,Bakas2017BrainlesionGM} is constructed for analysis and diagnosis of brain disease. The publicly available dataset of multi-institutional pre-operative MRI sequences is provided: training (1251 cases) and validation (219 cases). Each patient contributes 155$\times$240$\times$240 with four sequences: T1, T2, T1ce, and FLAIR. 

\textbf{Metrics.} We employ three metrics to evaluate our generator's performance. The first is the mean absolute error (MAE):
\begin{equation}\small
    MAE = \frac{1}{nm}\sum_{n}^{i=1}\sum_{m}^{j=1}\left| T_{ij} - G_{ij} \right|,
\end{equation}
where $T_{ij}$ denotes the ground truth neuroimage pixel and $G_{ij}$ denotes the generated neuroimage pixel. The lower value of MAE means the better performance.

The second metric is the peak signal-to-noise ratio (PSNR). PSNR is a function of the mean squared error and is better to evaluate the context (edge) detail of neuroimages. The higher PSNR value means the better performance. 
\begin{equation}\small
    PSNR = -10 \: log_{10}\left ( \frac{1}{nm}\sum_{n}^{i=1}\sum_{m}^{j=1} (T_{ij} - G_{ij})^{2} \right ).
\end{equation}
The third metric is structural similarity index (SSIM), which is a weighted combination of the luminance, the contrast and the structure. A higher SSIM value means the better performance.
\begin{equation}\small\label{eq:ssim}
\mathrm{SSIM}=\frac{\left(2 \mu_{T} \mu_{G}+C_{1}\right)\left(2 \sigma_{T G}+C_{2}\right)}{\left(\mu_{T}^{2}+\mu_{G}^{2}+C_{1}\right)\left(\sigma_{T}^{2}+\sigma_{G}^{2}+C_{2}\right)}.
\end{equation}
The $\mu$ and $\sigma$ in SSIM are the mean value and standard deviation of an image, respectively. $C_{1}$ and $C_{2}$ are two positive constants. We set $C_{1}$ and $C_{2}$ are 0.01 and 0.03, respectively.

\modified{\textbf{Federated Data Setting.}\label{sec:experiment_data_settings}
First, to ensure the validity and diversity of the data, we select 50 to 80 z-axis slices for each three-dimensional volume. The size of the data is uniformly cropped to 256 by 256 pixels. Second, to simulate data distributions in each client as closely as possible to real-world scenarios, we primarily divide the training data (volumes) proportionally among the predefined clients, where each client has its own private and unique ones. Using private data, we then construct paired and unpaired data based on the ratio specified for each client. Consequently, we can construct data patterns by adjusting the ratio between them (see Fig.~\ref{fig:data-spitted}).}

\modified{\textbf{Hyperparameter Setting.}
We use the learning rate of $1e-4$ and the batch size of 4. The optimizer is Adam~\citep{kingma2014adam}. Its beta1 and beta2 are 0.5 and 0.999, respectively. The weights of GAN loss $\lambda_{adv}$ and Cycle loss $\lambda_{cyc}$ are 1.0, 10.0, respectively. 
In FedMed-GAN differential privacy settings, the levels of gradient clip bound, sensitivity, and noise multiplier are fixed to 1.0, 2.0, and 1.07, respectively. In terms of the differential privacy setting, the Gaussian noise $\mu$ is set to 1.07, and the standard deviation $\sigma$ is set to 2.0. The clip bound for the back-propagation gradient is set to 1.0. }

\subsection{Analysis of FedMed-GAN}
From Table~\ref{ex:baseline-brats} and Table~\ref{ex:baseline-ixi}, we observe that FedMed-GAN does not sacrifice any performance under DP, and produces even better results compared with centralized training in almost all cases. \modified{In the experiment, we use 6000 images for both centralized and federated training. The ratio of paired data and unpaired data is fixed to 0.5. We set 30 epochs of centralized learning and 10 rounds of federated learning before aggregating the local models using FedAvg~\citep{mcmahan2017communication}, in which each client is trained with 3 epochs in one round. In federated scenario, the weights of client data are set to 0.4, 0.3, 0.2, and 0.1, respectively.}

\begin{table*}[ht]
	\centering 
	\caption{Baseline results on the BraTS2019 dataset.}
	\resizebox{0.95\textwidth}{!}{
    \begin{tabular}{c|c|c|c|c|c|c|c|c}
    \hline
                              & \textbf{Method} & \textbf{Indicator} & \textbf{T1 $\rightarrow$ T2} & \textbf{T1 $\rightarrow$ FLAIR} & \textbf{T2 $\rightarrow$ T1} & \textbf{T2 $\rightarrow$ FLAIR} & \textbf{FLAIR $\rightarrow$ T1} & \textbf{FLAIR $\rightarrow$ T2}\\ \hline
    \multirow{15}{*}{Non-Fed.} & \multirow{3}{*}{MUNIT}    & MAE~$\downarrow$  & 0.0452 & 0.0492 & 0.0466 & 0.0459 & \textbf{0.0382} & 0.0420 \\ \cline{3-9} 
                              &                           & PSNR~$\uparrow$ & \textbf{21.743} & 19.818& 20.123 & 20.141 & 21.623 & 21.721 \\ \cline{3-9} 
                              &                           & SSIM~$\uparrow$  & 0.8980 & 0.8853 & 0.9371 & 0.9023 & \textbf{0.9543} & 0.9117 \\ \cline{2-9} 
                              & \multirow{3}{*}{UNIT}     & MAE~$\downarrow$  & 0.0437 & 0.0507 & 0.0536 & 0.0487 & 0.0482 & \textbf{0.0410} \\ \cline{3-9} 
                              &                           & PSNR~$\uparrow$ & 20.833 & 20.190 & 20.067 & 20.266 & 20.493 & \textbf{21.212} \\ \cline{3-9} 
                              &                           & SSIM~$\uparrow$  & 0.8855 & 0.8936 & 0.9313 & 0.8930 & \textbf{0.9338} &  0.9101 \\ \cline{2-9} 
                              & \multirow{3}{*}{CycleGAN} & MAE~$\downarrow$  & 0.0567 & 0.0535 & 0.0557 & 0.0462 & 0.0471 & \textbf{0.0454} \\ \cline{3-9} 
                              &                           & PSNR~$\uparrow$ & 19.244 & 20.098 & 19.012 & 20.683 & 20.315 & \textbf{20.979} \\ \cline{3-9} 
                              &                           & SSIM~$\uparrow$  & 0.8427 & 0.8984 & 0.9192 & 0.9140 & \textbf{0.9371} & 0.9003 \\ \cline{2-9} 
                              & \multirow{3}{*}{\modified{HLHGAN}} & MAE~$\downarrow$  & \textbf{0.0217} & 0.0234 & 0.0245 & 0.0237 & 0.0246 & 0.0254 \\ \cline{3-9} 
                              &                           & PSNR~$\uparrow$ & 30.847 & 30.578 & \textbf{30.876} & 30.847 & 30.702 & 30.689 \\ \cline{3-9} 
                              &                           & SSIM~$\uparrow$  & 0.8127 & 0.8137 & 0.8145 & \textbf{0.8167} & 0.8130 & 0.8146\\ \cline{2-9} 
                              & \multirow{3}{*}{\modified{Hyper-GAN}} & MAE~$\downarrow$  & 0.0073 & \textbf{0.0071} & 0.0081 & 0.0083 & 0.0084 & 0.0082 \\ \cline{3-9} 
                              &                           & PSNR~$\uparrow$ & 31.805 & \textbf{31.901} & 30.894 & 30.521 & 30.456 & 30.745 \\ \cline{3-9} 
                              &                           & SSIM~$\uparrow$  & 0.9270 & \textbf{0.9310} & 0.9057 & 0.9032 & 0.9021 & 0.9128 \\ \hline
    \multirow{15}{*}{Fed.}     & \multirow{3}{*}{MUNIT}    & MAE~$\downarrow$  & 0.0783 & 0.0934 & 0.0972 & 0.0960 & 0.1031 & \textbf{0.0782} \\ \cline{3-9} 
                              &                           & PSNR~$\uparrow$ & 16.653 & 15.377  & 15.724 & 15.275  & 14.931 & \textbf{17.048} \\ \cline{3-9} 
                              &                           & SSIM~$\uparrow$  & 0.6850  & 0.6463 & \textbf{0.8325}  & 0.6402 & 0.8048 & 0.7114 \\ \cline{2-9} 
                              & \multirow{3}{*}{UNIT}     & MAE~$\downarrow$  & 0.2631 & 0.2458 & 0.3461 & \textbf{0.2331} & 0.3443 & 0.3150 \\ \cline{3-9} 
                              &                           & PSNR~$\uparrow$ & 9.3403 & 9.5427 & 7.9660 & \textbf{10.284} & 7.4321 & 7.6504 \\ \cline{3-9} 
                              &                           & SSIM~$\uparrow$  & 0.0563 & \textbf{0.2066} & 0.0359 & 0.1768 & 0.0187 & 0.0555 \\ \cline{2-9} 
                              & \multirow{3}{*}{CycleGAN} & MAE~$\downarrow$  & 0.0609 &  0.0565 & 0.0565  & 0.0508  & \textbf{0.0470}  & 0.0591 \\ \cline{3-9} 
                              &                           & PSNR~$\uparrow$ & 18.506 & 18.620 & 18.620 & 19.904 & \textbf{20.540} & 19.402 \\ \cline{3-9} 
                              &                           & SSIM~$\uparrow$  & 0.8507 & 0.9146 & 0.9146 & 0.8954 & \textbf{0.9405} & 0.8666 \\ \cline{2-9} 
                               & \multirow{3}{*}{\modified{HLHGAN}} & MAE~$\downarrow$  & \textbf{0.0225} &  0.0242 & 0.0256  & 0.0257  & 0.0249  & 0.0261 \\ \cline{3-9} 
                              &                           & PSNR~$\uparrow$ & \textbf{30.852} & 30.523 & 30.456 & 29.843 & 29.478 & 29.587 \\ \cline{3-9} 
                              &                           & SSIM~$\uparrow$  & 0.8032 & 0.8131 & 0.8121 & 0.8135 & 0.8042 & \textbf{0.8212} \\ \cline{2-9}  
                              & \multirow{3}{*}{\modified{Hyper-GAN}} & MAE~$\downarrow$  & 0.0082 &  \textbf{0.0081} & 0.0092  & 0.0085  & 0.0094  & 0.0086 \\ \cline{3-9} 
                              &                           & PSNR~$\uparrow$ & \textbf{31.905} & 31.575 & 29.894 & 30.323 & 30.578 & 30.812 \\ \cline{3-9} 
                              &                           & SSIM~$\uparrow$  & 0.9183 & \textbf{0.9237} & 0.9057 & 0.9032 & 0.9046 & 0.9217 \\ \hline 
    \end{tabular}
    }
    \label{ex:baseline-brats}
\end{table*}
\begin{table}[ht]
	\centering 
	\caption{Baseline results on the IXI dataset.}
	\resizebox{0.48\textwidth}{!}{
	\begin{tabular}{c|c|c|c|c}
    \hline
                              & \textbf{Method} & \textbf{Indicator} & \textbf{T2 $\rightarrow$ PD} & \textbf{PD $\rightarrow$ T2}  \\ \hline
    \multirow{15}{*}{Non-Fed.} & \multirow{3}{*}{MUNIT}    & MAE~$\downarrow$  & 0.0366 & \textbf{0.0324}\\ \cline{3-5} 
                              &                           & PSNR~$\uparrow$ & 23.498 & \textbf{23.994} \\ \cline{3-5} 
                              &                           & SSIM~$\uparrow$  & \textbf{0.9666} & 0.9524 \\ \cline{2-5} 
                              & \multirow{3}{*}{UNIT}     & MAE~$\downarrow$  & 0.0417 & \textbf{0.0356} \\ \cline{3-5} 
                              &                           & PSNR~$\uparrow$ & 22.671 & \textbf{23.000} \\ \cline{3-5} 
                              &                           & SSIM~$\uparrow$  & \textbf{0.9575} & 0.9380 \\ \cline{2-5} 
                              & \multirow{3}{*}{CycleGAN} & MAE~$\downarrow$  & 0.0356 & \textbf{0.0315} \\ \cline{3-5} 
                              &                           & PSNR~$\uparrow$ & \textbf{24.702} & 24.145 \\ \cline{3-5} 
                              &                           & SSIM~$\uparrow$  & \textbf{0.9715} & 0.9516 \\ \cline{2-5}
                              & \multirow{3}{*}{\modified{HLHGAN}} & MAE~$\downarrow$  & 0.0372 & \textbf{0.0352} \\ \cline{3-5} 
                              &                           & PSNR~$\uparrow$ & 30.045 & \textbf{30.176} \\ \cline{3-5} 
                              &                           & SSIM~$\uparrow$  & \textbf{0.8872} & 0.8745 \\ \cline{2-5}
                              & \multirow{3}{*}{\modified{Hyper-GAN}} & MAE~$\downarrow$  & \textbf{0.0137} & 0.0148 \\ \cline{3-5} 
                              &                           & PSNR~$\uparrow$ & \textbf{30.521} & 30.307 \\ \cline{3-5} 
                              &                           & SSIM~$\uparrow$  & 0.9117 & \textbf{0.9139} \\ \hline
    \multirow{15}{*}{Fed.}     & \multirow{3}{*}{MUNIT}    & MAE~$\downarrow$  & 0.1093 & \textbf{0.0842} \\ \cline{3-5} 
                              &                           & PSNR~$\uparrow$ & 16.263 & \textbf{17.057} \\ \cline{3-5} 
                              &                           & SSIM~$\uparrow$  & \textbf{0.8393} & 0.7033 \\ \cline{2-5} 
                              & \multirow{3}{*}{UNIT}     & MAE~$\downarrow$  & 0.9800 & \textbf{0.2570} \\ \cline{3-5} 
                              &                           & PSNR~$\uparrow$ & \textbf{9.4781} & 8.8979 \\ \cline{3-5} 
                              &                           & SSIM~$\uparrow$  & \textbf{0.1587} & 0.0471 \\ \cline{2-5} 
                              & \multirow{3}{*}{CycleGAN} & MAE~$\downarrow$  & 0.0401 & \textbf{0.0364} \\ \cline{3-5} 
                              &                           & PSNR~$\uparrow$ & \textbf{24.061} & 23.824 \\ \cline{3-5} 
                              &                           & SSIM~$\uparrow$  & \textbf{0.9666} & 0.9472 \\ \cline{2-5}
                              & \multirow{3}{*}{\modified{HLHGAN}} & MAE~$\downarrow$  & 0.0275 & \textbf{0.0273} \\ \cline{3-5} 
                              &                           & PSNR~$\uparrow$ & \textbf{30.031} & 29.852 \\ \cline{3-5} 
                              &                           & SSIM~$\uparrow$  & 0.8653 & \textbf{0.8677} \\ \cline{2-5}
                              & \multirow{3}{*}{\modified{Hyper-GAN}} & MAE~$\downarrow$  & \textbf{0.0132} & 0.0133 \\ \cline{3-5} 
                              &                           & PSNR~$\uparrow$ & \textbf{29.641} & 29.342 \\ \cline{3-5} 
                              &                           & SSIM~$\uparrow$  & \textbf{0.9103} & 0.9098 \\ \hline
    \end{tabular}
    }
    \label{ex:baseline-ixi}
\end{table}

\begin{figure}[ht]
    \centering
    \includegraphics[width=0.48\textwidth]{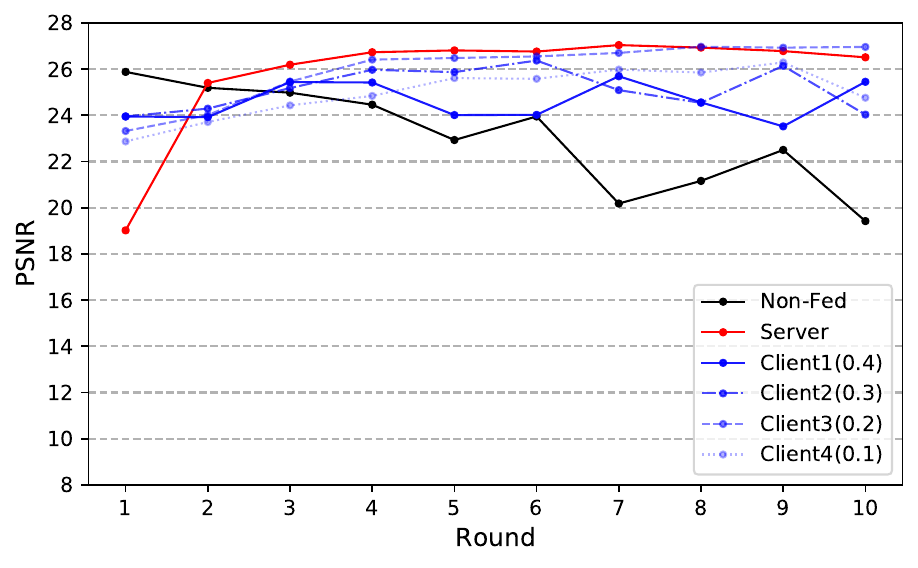}
	\caption{\modified{Results of FedMed-GAN training through multiple communication rounds.}}
 \label{fig:loss_analysis}
\end{figure}

Fig.~\ref{fig:loss_analysis} shows ablation studies of multiple communication rounds on the IXI dataset. Note that the red line indicates the performance of the server's generator in FedMed-GAN. The blue line denotes the performance of generator under the centralized training. We observe that FedMed-GAN is able to stabilize the training dynamics of the generator, and gradually increase the performance of the server's generator. However, the mode collapse problem occurs in the centralized training method. We also note that the performance of generator starts to deteriorate in the 21$th$ epoch from Fig.~\ref{fig:loss_analysis}.

\subsubsection{Multiple Clients and Data Distribution}
\modified{We examine the effectiveness of FedMed-GAN when confronted with long-tail data phenomena. Table~\ref{ex:data-distribution} demonstrates that FedMed-GAN is more stable.} In Table~\ref{ex:data-distribution}, we divide the data distribution method into [`average', `gradual', `extreme']. For instance, when the client number is 4 and the proportion is 0.7+0.1$\times$3 (extreme mode). 
\modified{It indicates that the original data assign 70\% to the first client and 10\% to each of the remaining three clients.}

\begin{figure*}[thb]
	\centering
    \includegraphics[width=1\textwidth]{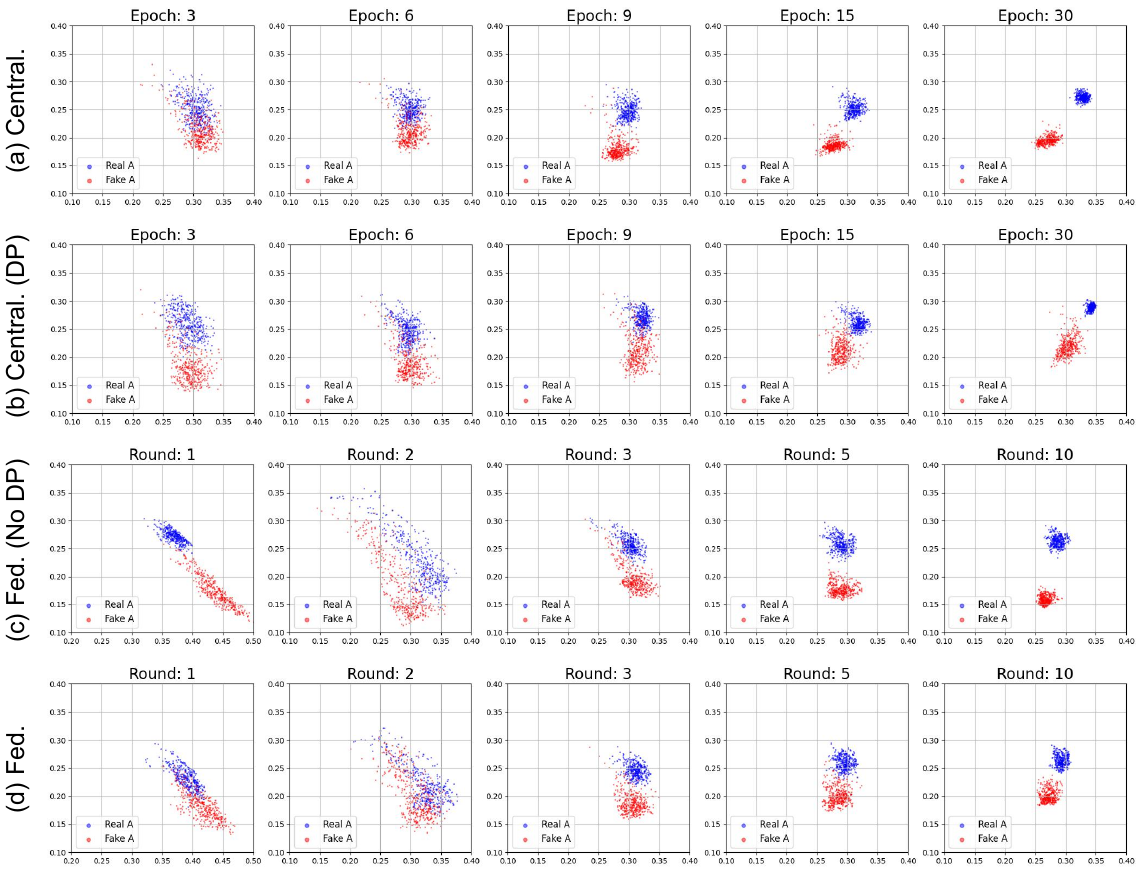}
	\caption{
		Visualization of sample's features, which are generated by FedMed-GAN and centralized training.
	}\label{fig:latent-space-ixi}
\end{figure*}

\begin{table*}[ht]
	\centering 
	\caption{Results of FedMed-GAN on IXI (PD $\rightarrow$ T2).}
	\resizebox{0.7\textwidth}{!}{
    \begin{tabular}{c|c|c|c|c}
    \hline
    \textbf{Scheme}                   & \textbf{Client Num.}     & \textbf{2}         & \textbf{4}                     & \textbf{8}                           \\ \hline
    \multirow{4}{*}{Average} & Proportion & 0.5 $\times$ 2   & 0.25 $\times$ 4              & 0.125 $\times$ 8                      \\ \cline{2-5} 
                             & MAE~$\downarrow$       & 0.0258   & \textbf{0.0199}   &  0.0219   \\ \cline{2-5} 
                             & PSNR~$\uparrow$       & 26.9717   & \textbf{28.0443}   & 27.4549      \\ \cline{2-5} 
                             & SSIM $\uparrow$        & 0.9736    & \textbf{0.9785}   &  0.9775           \\ \hline
    \multirow{4}{*}{Gradual} & Proportion & 0.6 + 0.4 & \makecell[c]{0.4 + 0.3 +  0.2 + 0.1 } & \makecell[c]{0.3 + 0.2 + 0.1 $\times$ 4 \\+ 0.05 $\times$ 2} \\ \cline{2-5} 
                             & MAE~$\downarrow$       & 0.0252   &  \textbf{0.0203}  & 0.0204      \\ \cline{2-5} 
                             & PSNR~$\uparrow$      & 27.4915   & 27.7578   &  \textbf{27.8672}    \\ \cline{2-5} 
                             & SSIM $\uparrow$        & 0.9776    &  \textbf{0.9787}  & \textbf{0.9787}           \\ \hline
    \multirow{4}{*}{Extreme} & Proportion & 0.9 + 0.1   & 0.7 + 0.1 $\times$ 3     & 0.3 + 0.1 $\times$ 7                  \\ \cline{2-5} 
                             & MAE~$\downarrow$       & 0.0264   & 0.0222   & \textbf{0.0218}     \\ \cline{2-5} 
                             & PSNR~$\uparrow$         & 26.8523  & 27.5999   & \textbf{28.1727}          \\ \cline{2-5} 
                             & SSIM $\uparrow$        & 0.9737   & 0.9783   &  \textbf{0.9889}                         \\ \hline
    \end{tabular}
    }
    \label{ex:data-distribution}
\end{table*}

\subsubsection{Latent Space of Generators}
To deeply investigate why FedMed-GAN can outperform the centralized training, we visualize the latent space of generators for FedMed-GAN and centralized training, respectively. In specific, the generator of CycleGAN is U-Net~\citep{Ronneberger2015UNetCN}. We take the down-sample layer 5 of U-Net as the latent space. After that, we divide the last dimension of the extracted vector into two dimensions (denoted as x and y), and reduce the vector by averaging the remaining dimensions. In Fig.~\ref{fig:latent-space-ixi}, we plot 400 samples randomly selected in each modality.

The main differences between our FedMed-GAN and the centralized training are FedAvg~\citep{mcmahan2017communication} and DP-SGD~\citep{abadi2016deep}. We provide ablation studies for these two parts and evaluate whether they play an important role in stabilizing the training dynamics of GAN. The aim of cross-modality neuroimaging data is to pursue high fidelity and global diversity. \modified{The global diversity refers to the variety of neuroimaging data points as a whole. The cross-modality natural image synthesis attempts to synthesize each image into multiple styles as opposed to a single style.} As for fidelity, we can obtain the results from the overlapped size of the latent space between the real ones. As the overlapped size becomes larger, the fidelity of the generated samples becomes higher. For global diversity, we can obtain the results from the variance of latent space of the generated samples, i.e. fake A and fake B. Therefore, as the variance becomes higher, the global diversity gets much higher.

\subsubsection{Notation}
In Fig.~\ref{fig:latent-space-ixi}, modal A is PD-weighted MRI data and modal B is T2-weighted MRI data from IXI. The red dots denote the distributions of the latent space for modal A. The blue dots are the distributions of the latent space for the generated samples of modal A~(termed as fake A). The green dots are the distributions of the latent space for modal B. The black dots are the distributions of the latent space for the generated samples of modal B~(termed as fake B). Fig.~\ref{fig:latent-space-ixi}(a) denotes the centralized training method for CycleGAN. Fig.~\ref{fig:latent-space-ixi}(b) denotes the centralized training method combined with DP-SGD. Fig.~\ref{fig:latent-space-ixi}(c) denotes the FedAvg algorithm without DP-SGD. Fig.~\ref{fig:latent-space-ixi}(d) denotes the FedAvg algorithm combined with DP-SGD.

\subsubsection{Differential Privacy}\label{sec:dp}

\textit{(1) FedAvg+DPSGD vs FedAvg?}
Firstly, we investigate the role of DP. When comparing Fig.~\ref{fig:latent-space-ixi}(c) and Fig.~\ref{fig:latent-space-ixi}(d), we observe that FedAvg+DPSGD outperforms FedAvg. \modified{Also, the diversity of fake A in FedAvg+DP-SGD is significantly greater than in FedAvg without DPSGD. In addition, from round 5 to round 10, the distance between the real A and the fake A in FedAvg+DPSGD decreases, whereas the latent space between the real A and the fake A in FedAvg without DP-SGD gradually increases. It suggests that DP-SGD stabilizes the training dynamics of GAN.}

\begin{table}[ht]
	\centering 
	\caption{Noise comparisons on the IXI dataset.}
	\resizebox{0.45\textwidth}{!}{
        \begin{tabular}{c|c|c|c|c}
        \hline
         \textbf{Noise} & \textbf{\textbf{Clip-Bound}}                        & \makecell[c]{ \textbf{0.7}} &  \makecell[c]{\textbf{1.0}}  &  \makecell[c]{\textbf{1.3}}  \\ \hline
        \multirow{3}{*}{\makecell[c]{0.5}}       & MAE~~$\downarrow$  &  \textbf{0.0198}      & 0.0200     &  0.0226     \\ \cline{2-5} 
                                                              & PSNR $\uparrow$ & 27.8496      & \textbf{27.934}       &  27.8617           \\ \cline{2-5} 
                                                          & SSIM $\uparrow$ &  0.9787   & \textbf{0.9794}      & 0.9786       \\ \hline
      \multirow{3}{*}{\makecell[c]{1.07}}       & MAE~~$\downarrow$  &  0.0212      &  0.0203      &  \textbf{0.0201}           \\ \cline{2-5} 
                                                          & PSNR $\uparrow$ & \textbf{28.0310}      & 27.7578       &  27.7840       \\ \cline{2-5} 
                                                          & SSIM $\uparrow$ & \textbf{0.9801}      &  0.9787      &  0.9786         \\ \hline
      \multirow{3}{*}{\makecell[c]{2.0}}        & MAE~~$\downarrow$  &  \textbf{0.0203}    &  0.0204      & 0.0222          \\ \cline{2-5} 
                                                          & PSNR $\uparrow$ & \textbf{27.8385}     &  27.8125      & 27.8149     \\ \cline{2-5} 
                                                          & SSIM $\uparrow$ & 0.9784    & \textbf{0.9786}       &   0.9782     \\ \hline
        \end{tabular}
        }
    \label{ex:dp-noise}
\end{table} 

\begin{table}[!ht]
    \centering
    \caption{Centralized training with more discriminators on the IXI dataset. D denotes the number of the discriminators. Each grid's value is the PSNR score.}
    \begin{tabular}{c|c|c|c}
    \hline
        \textbf{Epoch} & \textbf{D=2} & \textbf{D=4} & \textbf{D=8} \\ \hline
        1 & \textbf{25.1043} & \textbf{24.8932} & \textbf{24.9375} \\ \hline
        6 & 24.1453 & 23.9843 & 23.7685 \\ \hline
        9 & 23.6572 & 23.3483 & 22.5643 \\ \hline
        15 & 21.4587 & 21.3987 & 20.8761 \\ \hline
        18 & 20.8875 & 20.8846 & 20.1698 \\ \hline
        24 & 19.2497 & 19.1430 & 18.9456 \\ \hline
        30 & 18.2768 & 18.3674 & 18.3654 \\ \hline
    \end{tabular}
    \label{tab:ixi_centralized_discriminators}
\end{table}

\begin{table}[!ht]
    \centering
    \caption{Centralized training with more discriminators on the BraTS dataset. D denotes the number of discriminators. Each grid's value is the PSNR score.}
    \begin{tabular}{c|c|c|c}
    \hline
        \textbf{Epoch} & \textbf{D=2} & \textbf{D=4} & \textbf{D=8} \\ \hline
        1 & 18.9329 & 18.2671 & 18.1451 \\ \hline
        6 & \textbf{19.3435} & \textbf{18.9982} & \textbf{18.9213} \\ \hline
        9 & 18.3467 & 18.1656 & 18.1345 \\ \hline
        15 & 17.6589 & 17.4532 & 17.2869 \\ \hline
        18 & 16.6945 & 16.5672 & 16.4783 \\ \hline
        24 & 16.2497 & 15.9843 & 15.7566 \\ \hline
        30 & 15.9345 & 15.6756 & 15.6783 \\ \hline
    \end{tabular}
    \label{tab:brats_centralized_discriminators}
\end{table}

\textit{(2) Central+DPSGD vs Central?} Furthermore, we add DP-SGD into the centralized training, as described in Fig.~\ref{fig:latent-space-ixi}(b). Although adding DP-SGD into the centralized training is unnecessary, we desire to verify our assumption that DP-SGD can be treated as a gradient penalty to prevent mode collapse. Compared with Fig.~\ref{fig:latent-space-ixi}(a) and Fig.~\ref{fig:latent-space-ixi}(b), we can find that the global diversity of fake A is much larger in centralized training+~DP-SGD than the ones in centralized training in the absence of DP-SGD after epoch 12. The distance between fake A and real A becomes larger in centralized training without DP-SGD after epoch 9. 
However, the distribution of fake A and real A is much farther after epoch 9. We think that DP-SGD can be one of the important measures to facilitate the convergence of GAN. Rethinking the Eq. (1) to Eq. (3), we find that DP-SGD works as a gradient penalty, which has been proved as one of simple but effective approach to regularize the training dynamics of GAN~\citep{Mescheder2018WhichTM}. Such a result has concurred with each other in this view. Finally, to evaluate the hyperparameters of clip bound and noise density in DP, we perform relevant experiments and find that the performance of FedMed-GAN is very stable with various DP guarantees. These results are provided in Table~\ref{ex:dp-noise}.

\subsubsection{FedAvg}
\modified{Despite DP-SGD, FedAvg in Algorithm~\ref{alg:server_g} also plays an important role in stabilizing the training.} Compared with Fig.~\ref{fig:latent-space-ixi}(a) and Fig.~\ref{fig:latent-space-ixi}(d), the distance between real A samples and fake A samples is gradually smaller in FedAvg after epoch 9. However, the distributions of real A and fake A are gradually distant in the centralized training after epoch 9. The Fed-Avg method aggregates the weight from each client's generator and edge network to the server model according to the data proportion distribution. The manifolds of the generators (the server's generators and the clients' ones) in FedMed-GAN are reparamterized, which are consistent with the assumption of convergence proof in Mescheder~\textit{et al.}~\citep{Mescheder2018WhichTM}.

\begin{table*}[htbp]
    \centering
    \caption{Centralized Training + DP-SGD. 'w' denotes the centralized training with DP-SGD. 'w/o' denotes the centralized training without DP-SGD. Each grid's value is the PSNR score.}
    \resizebox{0.85\textwidth}{!}{
    \begin{tabular}{c|c|c|c|c|c|c|c|c|c|c|c|c}
    \hline
        \textbf{Dataset} & \textbf{Epoch} & \textbf{1} & \textbf{4} & \textbf{7} & \textbf{10} & \textbf{13} & \textbf{16} & \textbf{19} & \textbf{21} & \textbf{24} & \textbf{27} & \textbf{30} \\ \hline
        IXI & w/o & 23.69 & 25.03 & 24.10 & \textbf{25.30} & 24.91 & 23.11 & 23.49 & 20.18 & 21.16 & 22.50 & 19.42 \\ \cline{2-13}
        (PD-T2) & w & 25.67 & 25.67 & \textbf{25.98} & 24.23 & 23.41 & 20.68 & 21.07 & 23.69 & 18.91 & 20.71 & 21.79 \\ \hline
        BraTS & w/o & 18.20 & 18.28 & 18.67 & 19.40 & 18.78 & 18.47 & 17.39 & 17.74 & \textbf{19.47} & 17.51 & 18.47 \\ \cline{2-13}
        (T1-FLAIR) & w & 19.34 & \textbf{19.94} & 19.20 & 19.43 & 19.53 & 19.50 & 19.53 & 19.40 & 18.95 & 19.61 & 16.82 \\ \hline
    \end{tabular}
    }
    \label{tab:centralized+dp}
\end{table*}

\subsubsection{Centralized+More Discriminators}

We add more discriminators in the centralized training experiment and the results are listed in Table~\ref{tab:ixi_centralized_discriminators} and Table~\ref{tab:brats_centralized_discriminators}. The number of discriminators is the same as the number of clients in federated learning. We average the logits from multiple discriminators for the training and testing. The results are provided here, which show that adding more discriminators is not able to stabilize the training dynamics. We think that the largest difference between centralized training and FedMed-GAN is a FedAvg algorithm. In addition, in the original paper, we have illustrated that the role of FedAvg is to~\citep{Mescheder2018WhichTM}.

\subsubsection{Centralized+DPSGD vs DPSGD}

We conduct centralized training + DP-SGD on IXI (PD$\rightarrow$T2) dataset and BraTS (T1$\rightarrow$FLAIR) dataset. The results are listed in Table~\ref{tab:centralized+dp}. We observe that centralized training + DP-SGD is not able to prevent mode collapse issue. We think that DP-SGD equals to gradient clipping and adding Gaussian noise to the clipped gradient. The discriminator overfitting problem is the main reason for mode collapse issue. The discriminator over-fitting can result in the gradient vanishing problem (i.e., the gradient is very small). In this case, the gradient clipping mechanism is meaningless since the gradient norm is much smaller than the clip bound. We observe that the gradient norms in centralized training approximately reach to 0 when mode collapse happens. Thus, DP-SGD + Centralized Training cannot mitigate mode collapse issue. Instead, we observe that the gradient norms in centralized training approximately reach to 0 when mode collapse happen issue. The reason is that the data assigned for each client's discriminator is limited. It is not easy to result in discriminator's over-fitting and not easy to be stuck into the local optimum. Therefore, DP-SGD can help mitigate the mode collapse issue in FedMed-GAN.

\begin{table*}[ht]
    \centering
    \renewcommand{\arraystretch}{1.3}
    \caption{PSNR's performance on various differential privacy budgets. In each grid, the first term is the privacy budget and the PSNR score performance $p$. $\delta$ denotes the standard deviation of Gaussian noise. $\epsilon$ is the privacy budget indicating the privacy level. The smaller $\epsilon$ value implies stronger privacy protection. }
    \resizebox{0.75\textwidth}{!}{
    \begin{tabular}{c|c|c|c}
    \hline
        \textbf{($\epsilon$, $\delta$)} & \textbf{2 clients}  & \textbf{4 clients} & \textbf{8 clients}\\ \hline
        noise=0.5 & (813.79, 1e-5), $p$: 27.4678 & (505.26, 1e-5), $p$: 28.0123 & (250.44, 1e-5), $p$=27.1564 \\ \hline
        noise=1.07 & (200.47, 1e-5) $p$: 27.7341 & (74.36, 1e-05), $p$: 28.1374 & (28.83, 1e-5), $p$=27.1432 \\ \hline
        noise=2.0 & (71.51, 1e-5) $p$: 27.8921 & (27.98, 1e-5), $p$: 28.1475 & (12.92, 1e-5), $p$=27.1876 \\ \hline
    \end{tabular}
    }
    \label{tab:pb}
\end{table*}

\begin{figure*}[htb]
    \centering 
  \includegraphics[width=0.85\textwidth]{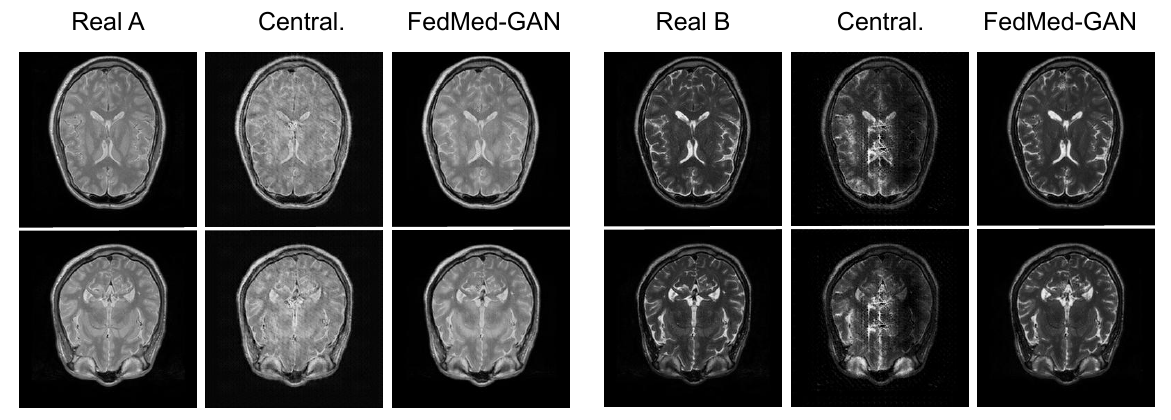}
\caption{Visualization of FedMed-GAN. We use the results of 24$th$ epoch of centralized training (Central.) and 8$th$ round of FedMed-GAN as examples (IXI, PD $\leftrightarrow$ T2).}
\label{fig:vis-ixi}
\end{figure*}

\subsubsection{Privacy Budget}

According to Eq.~(\ref{eq:privacy_budget}), we calculate the privacy budget with various Gaussian noises. From Table~\ref{tab:pb}, we find that the synthesis quality of FedMed-GAN is not affected by the different privacy budgets. It also indicates the robustness of FedMed-GAN.

\subsubsection{Noise Comparisons in DP}

Moreover, we desire to know whether the effect resulted from DP-SGD are very sensitive to the hyper-parameter setting or not. So if yes, it may violate our explanations mentioned above. But from Table~\ref{ex:dp-noise}, we find that the performance of FedMed-GAN is very stable with various DP guarantees.

\section{Visualization}
Fig.~\ref{fig:vis-ixi} presents visualization resulting achieved by FedMed-GAN. Real A (PD) and Real B (T2) are the ground-truth neuroimage. Cetralized denotes the synthesis neuroimage via centralized training. FedMed-GAN denotes the synthesis neuroimage by FedMed-GAN. From Fig.~\ref{fig:vis-ixi}, we can observe that FedMed-GAN can generate high-quality images with vivid information of brain tissues, compared with ones generated by centralized training.

\section{Conclusion}

In this paper, we propose a novel baseline FedMed-GAN for unsupervised cross-modality brain image synthesis in federated manner. We find that both DP and the federated averaging mechanism can effectively improve the ability of the model generation, showing advantages over the centralized training. Experimental results show that FedMed-GAN can generate clear images, providing a powerful model for the field of federated domain translation. \textbf{Limitations and Future Work: } The performance of FedMed-GAN has not been verified by the radiologist and the interpretability needs to be explored in the future.



\printcredits

\section*{Declaration of Competing Interest} 
The authors declare that they have no known competing financial interests or personal relationships that could have appeared to influence the work reported in this paper.

\section*{Acknowledgments} 
This work was supported by the National Key R\&D Program of China (Grant NO. 2022YFF1202903) and the National Natural Science Foundation of China (Grant NO. 62122035,  62206122 and 61972188). Y. Jin is supported by an Alexander von Humboldt Professorship for AI endowed by the German Federal Ministry of Education and Research.

\bibliographystyle{cas-model2-names}

\bibliography{cas-refs}

\begin{thebibliography}{38}
\expandafter\ifx\csname natexlab\endcsname\relax\def\natexlab#1{#1}\fi
\providecommand{\url}[1]{\texttt{#1}}
\providecommand{\href}[2]{#2}
\providecommand{\path}[1]{#1}
\providecommand{\DOIprefix}{doi:}
\providecommand{\ArXivprefix}{arXiv:}
\providecommand{\URLprefix}{URL: }
\providecommand{\Pubmedprefix}{pmid:}
\providecommand{\doi}[1]{\href{http://dx.doi.org/#1}{\path{#1}}}
\providecommand{\Pubmed}[1]{\href{pmid:#1}{\path{#1}}}
\providecommand{\bibinfo}[2]{#2}
\ifx\xfnm\relax \def\xfnm[#1]{\unskip,\space#1}\fi
\bibitem[{Abadi et~al.(2016)Abadi, Chu, Goodfellow, McMahan, Mironov, Talwar
  and Zhang}]{abadi2016deep}
\bibinfo{author}{Abadi, M.}, \bibinfo{author}{Chu, A.},
  \bibinfo{author}{Goodfellow, I.}, \bibinfo{author}{McMahan, H.B.},
  \bibinfo{author}{Mironov, I.}, \bibinfo{author}{Talwar, K.},
  \bibinfo{author}{Zhang, L.}, \bibinfo{year}{2016}.
\newblock \bibinfo{title}{Deep learning with differential privacy}, in:
  \bibinfo{booktitle}{Proceedings of the 2016 ACM SIGSAC conference on computer
  and communications security}, pp. \bibinfo{pages}{308--318}.
\bibitem[{Aljabar et~al.(2011)Aljabar, Wolz, Srinivasan, Counsell, Rutherford,
  Edwards, Hajnal and Rueckert}]{Aljabar2011ACM}
\bibinfo{author}{Aljabar, P.}, \bibinfo{author}{Wolz, R.},
  \bibinfo{author}{Srinivasan, L.}, \bibinfo{author}{Counsell, S.J.},
  \bibinfo{author}{Rutherford, M.A.}, \bibinfo{author}{Edwards, A.D.},
  \bibinfo{author}{Hajnal, J.V.}, \bibinfo{author}{Rueckert, D.},
  \bibinfo{year}{2011}.
\newblock \bibinfo{title}{A combined manifold learning analysis of shape and
  appearance to characterize neonatal brain development}.
\newblock \bibinfo{journal}{IEEE Transactions on Medical Imaging}
  \bibinfo{volume}{30}, \bibinfo{pages}{2072--2086}.
\bibitem[{Augenstein et~al.(2020)Augenstein, McMahan, Ramage, Ramaswamy,
  Kairouz, Chen, Mathews and y~Arcas}]{DBLP:conf/iclr/AugensteinMRRKC20}
\bibinfo{author}{Augenstein, S.}, \bibinfo{author}{McMahan, H.B.},
  \bibinfo{author}{Ramage, D.}, \bibinfo{author}{Ramaswamy, S.},
  \bibinfo{author}{Kairouz, P.}, \bibinfo{author}{Chen, M.},
  \bibinfo{author}{Mathews, R.}, \bibinfo{author}{y~Arcas, B.A.},
  \bibinfo{year}{2020}.
\newblock \bibinfo{title}{Generative models for effective {ML} on private,
  decentralized datasets}, in: \bibinfo{booktitle}{{ICLR}},
  \bibinfo{publisher}{OpenReview.net}.
\bibitem[{Bakas et~al.(2017)Bakas, Kuijf, Menze and
  Reyes}]{Bakas2017BrainlesionGM}
\bibinfo{author}{Bakas, S.}, \bibinfo{author}{Kuijf, H.J.},
  \bibinfo{author}{Menze, B.H.}, \bibinfo{author}{Reyes, M.},
  \bibinfo{year}{2017}.
\newblock \bibinfo{title}{Brainlesion: Glioma, multiple sclerosis, stroke and
  traumatic brain injuries}, in: \bibinfo{booktitle}{Lecture Notes in Computer
  Science}.
\bibitem[{Chen et~al.(2020a)Chen, Orekondy and Fritz}]{chen2020gs}
\bibinfo{author}{Chen, D.}, \bibinfo{author}{Orekondy, T.},
  \bibinfo{author}{Fritz, M.}, \bibinfo{year}{2020}a.
\newblock \bibinfo{title}{Gs-wgan: A gradient-sanitized approach for learning
  differentially private generators}.
\newblock \bibinfo{journal}{Advances in Neural Information Processing Systems}
  \bibinfo{volume}{33}, \bibinfo{pages}{12673--12684}.
\bibitem[{Chen et~al.(2020b)Chen, Orekondy and Fritz}]{DBLP:conf/nips/ChenOF20}
\bibinfo{author}{Chen, D.}, \bibinfo{author}{Orekondy, T.},
  \bibinfo{author}{Fritz, M.}, \bibinfo{year}{2020}b.
\newblock \bibinfo{title}{{GS-WGAN:} {A} gradient-sanitized approach for
  learning differentially private generators}, in:
  \bibinfo{booktitle}{NeurIPS}.
\bibitem[{Dar et~al.(2019)Dar, Yurt, Karacan, Erdem, Erdem and
  Çukur}]{Dar2019ImageSI}
\bibinfo{author}{Dar, S.U.H.}, \bibinfo{author}{Yurt, M.},
  \bibinfo{author}{Karacan, L.}, \bibinfo{author}{Erdem, A.},
  \bibinfo{author}{Erdem, E.}, \bibinfo{author}{Çukur, T.},
  \bibinfo{year}{2019}.
\newblock \bibinfo{title}{Image synthesis in multi-contrast mri with
  conditional generative adversarial networks}.
\newblock \bibinfo{journal}{IEEE Transactions on Medical Imaging}
  \bibinfo{volume}{38}, \bibinfo{pages}{2375--2388}.
\bibitem[{Dwork and Roth(2014)}]{Dwork2014TheAF}
\bibinfo{author}{Dwork, C.}, \bibinfo{author}{Roth, A.}, \bibinfo{year}{2014}.
\newblock \bibinfo{title}{The algorithmic foundations of differential privacy}.
\newblock \bibinfo{journal}{Found. Trends Theor. Comput. Sci.}
  \bibinfo{volume}{9}, \bibinfo{pages}{211--407}.
\bibitem[{Goodfellow et~al.(2014)Goodfellow, Pouget-Abadie, Mirza, Xu,
  Warde-Farley, Ozair, Courville and Bengio}]{goodfellow2014generative}
\bibinfo{author}{Goodfellow, I.}, \bibinfo{author}{Pouget-Abadie, J.},
  \bibinfo{author}{Mirza, M.}, \bibinfo{author}{Xu, B.},
  \bibinfo{author}{Warde-Farley, D.}, \bibinfo{author}{Ozair, S.},
  \bibinfo{author}{Courville, A.}, \bibinfo{author}{Bengio, Y.},
  \bibinfo{year}{2014}.
\newblock \bibinfo{title}{Generative adversarial nets}.
\newblock \bibinfo{journal}{Advances in neural information processing systems}
  \bibinfo{volume}{27}.
\bibitem[{Guo et~al.(2021)Guo, Wang, Yasarla, Zhou, Patel and
  Jiang}]{Guo2021AnatomicAM}
\bibinfo{author}{Guo, P.}, \bibinfo{author}{Wang, P.},
  \bibinfo{author}{Yasarla, R.}, \bibinfo{author}{Zhou, J.},
  \bibinfo{author}{Patel, V.M.}, \bibinfo{author}{Jiang, S.},
  \bibinfo{year}{2021}.
\newblock \bibinfo{title}{Anatomic and molecular mr image synthesis using
  confidence guided cnns}.
\newblock \bibinfo{journal}{IEEE Transactions on Medical Imaging}
  \bibinfo{volume}{40}, \bibinfo{pages}{2832--2844}.
\bibitem[{Huang et~al.(2018)Huang, Liu, Belongie and
  Kautz}]{huang2018multimodal}
\bibinfo{author}{Huang, X.}, \bibinfo{author}{Liu, M.Y.},
  \bibinfo{author}{Belongie, S.}, \bibinfo{author}{Kautz, J.},
  \bibinfo{year}{2018}.
\newblock \bibinfo{title}{Multimodal unsupervised image-to-image translation},
  in: \bibinfo{booktitle}{Proceedings of the European conference on computer
  vision (ECCV)}, pp. \bibinfo{pages}{172--189}.
\bibitem[{Huang et~al.(2020a)Huang, Zheng, Cong, Huang, Scott and
  Shao}]{Huang2020MCMTGANMC}
\bibinfo{author}{Huang, Y.}, \bibinfo{author}{Zheng, F.},
  \bibinfo{author}{Cong, R.}, \bibinfo{author}{Huang, W.},
  \bibinfo{author}{Scott, M.R.}, \bibinfo{author}{Shao, L.},
  \bibinfo{year}{2020}a.
\newblock \bibinfo{title}{Mcmt-gan: Multi-task coherent modality transferable
  gan for 3d brain image synthesis}.
\newblock \bibinfo{journal}{IEEE Transactions on Image Processing}
  \bibinfo{volume}{29}, \bibinfo{pages}{8187--8198}.
\bibitem[{Huang et~al.(2020b)Huang, Zheng, Wang, Jiang, Wang and
  Shao}]{Huang2020SuperResolutionAI}
\bibinfo{author}{Huang, Y.}, \bibinfo{author}{Zheng, F.},
  \bibinfo{author}{Wang, D.}, \bibinfo{author}{Jiang, J.},
  \bibinfo{author}{Wang, X.}, \bibinfo{author}{Shao, L.},
  \bibinfo{year}{2020}b.
\newblock \bibinfo{title}{Super-resolution and inpainting with degraded and
  upgraded generative adversarial networks}, in: \bibinfo{booktitle}{IJCAI}.
\bibitem[{Jiang et~al.(2019)Jiang, Lu, Wei and Xu}]{jiang2019synthesize}
\bibinfo{author}{Jiang, G.}, \bibinfo{author}{Lu, Y.}, \bibinfo{author}{Wei,
  J.}, \bibinfo{author}{Xu, Y.}, \bibinfo{year}{2019}.
\newblock \bibinfo{title}{Synthesize mammogram from digital breast
  tomosynthesis with gradient guided cgans}, in:
  \bibinfo{booktitle}{International Conference on Medical Image Computing and
  Computer-Assisted Intervention}, \bibinfo{organization}{Springer}. pp.
  \bibinfo{pages}{801--809}.
\bibitem[{Kong et~al.(2021)Kong, Lian, Huang, Hu, Zhou
  et~al.}]{kong2021breaking}
\bibinfo{author}{Kong, L.}, \bibinfo{author}{Lian, C.}, \bibinfo{author}{Huang,
  D.}, \bibinfo{author}{Hu, Y.}, \bibinfo{author}{Zhou, Q.}, et~al.,
  \bibinfo{year}{2021}.
\newblock \bibinfo{title}{Breaking the dilemma of medical image-to-image
  translation}.
\newblock \bibinfo{journal}{Advances in Neural Information Processing Systems}
  \bibinfo{volume}{34}, \bibinfo{pages}{1964--1978}.
\bibitem[{Li et~al.(2022)Li, Wu, Wang, Mao, Alsaadi and Zeng}]{Li2022AGF}
\bibinfo{author}{Li, H.}, \bibinfo{author}{Wu, P.}, \bibinfo{author}{Wang, Z.},
  \bibinfo{author}{Mao, J.F.}, \bibinfo{author}{Alsaadi, F.E.},
  \bibinfo{author}{Zeng, N.}, \bibinfo{year}{2022}.
\newblock \bibinfo{title}{A generalized framework of feature learning enhanced
  convolutional neural network for pathology-image-oriented cancer diagnosis}.
\newblock \bibinfo{journal}{Computers in biology and medicine}
  \bibinfo{volume}{151 Pt A}, \bibinfo{pages}{106265}.
\bibitem[{Li et~al.(2020)Li, Sahu, Zaheer, Sanjabi, Talwalkar and
  Smith}]{DBLP:conf/mlsys/LiSZSTS20}
\bibinfo{author}{Li, T.}, \bibinfo{author}{Sahu, A.K.},
  \bibinfo{author}{Zaheer, M.}, \bibinfo{author}{Sanjabi, M.},
  \bibinfo{author}{Talwalkar, A.}, \bibinfo{author}{Smith, V.},
  \bibinfo{year}{2020}.
\newblock \bibinfo{title}{Federated optimization in heterogeneous networks},
  in: \bibinfo{booktitle}{MLSys}, \bibinfo{publisher}{mlsys.org}.
\bibitem[{Liu et~al.(2017)Liu, Breuel and Kautz}]{DBLP:conf/nips/LiuBK17}
\bibinfo{author}{Liu, M.}, \bibinfo{author}{Breuel, T.M.},
  \bibinfo{author}{Kautz, J.}, \bibinfo{year}{2017}.
\newblock \bibinfo{title}{Unsupervised image-to-image translation networks},
  in: \bibinfo{booktitle}{{NIPS}}, pp. \bibinfo{pages}{700--708}.
\bibitem[{Long et~al.(2019)Long, Wang, Yang, Kailkhura, Zhang, Gunter and
  Li}]{Long2019GPATESD}
\bibinfo{author}{Long, Y.}, \bibinfo{author}{Wang, B.}, \bibinfo{author}{Yang,
  Z.}, \bibinfo{author}{Kailkhura, B.}, \bibinfo{author}{Zhang, A.},
  \bibinfo{author}{Gunter, C.}, \bibinfo{author}{Li, B.}, \bibinfo{year}{2019}.
\newblock \bibinfo{title}{G-pate: Scalable differentially private data
  generator via private aggregation of teacher discriminators}, in:
  \bibinfo{booktitle}{Neural Information Processing Systems}.
\bibitem[{McMahan et~al.(2017a)McMahan, Moore, Ramage, Hampson and
  y~Arcas}]{mcmahan2017communication}
\bibinfo{author}{McMahan, B.}, \bibinfo{author}{Moore, E.},
  \bibinfo{author}{Ramage, D.}, \bibinfo{author}{Hampson, S.},
  \bibinfo{author}{y~Arcas, B.A.}, \bibinfo{year}{2017}a.
\newblock \bibinfo{title}{Communication-efficient learning of deep networks
  from decentralized data}, in: \bibinfo{booktitle}{Artificial intelligence and
  statistics}, \bibinfo{organization}{PMLR}. pp. \bibinfo{pages}{1273--1282}.
\bibitem[{McMahan et~al.(2017b)McMahan, Moore, Ramage, Hampson and
  y~Arcas}]{DBLP:conf/aistats/McMahanMRHA17}
\bibinfo{author}{McMahan, B.}, \bibinfo{author}{Moore, E.},
  \bibinfo{author}{Ramage, D.}, \bibinfo{author}{Hampson, S.},
  \bibinfo{author}{y~Arcas, B.A.}, \bibinfo{year}{2017}b.
\newblock \bibinfo{title}{Communication-efficient learning of deep networks
  from decentralized data}, in: \bibinfo{booktitle}{{AISTATS}},
  \bibinfo{publisher}{{PMLR}}. pp. \bibinfo{pages}{1273--1282}.
\bibitem[{Mescheder et~al.(2018)Mescheder, Geiger and
  Nowozin}]{Mescheder2018WhichTM}
\bibinfo{author}{Mescheder, L.M.}, \bibinfo{author}{Geiger, A.},
  \bibinfo{author}{Nowozin, S.}, \bibinfo{year}{2018}.
\newblock \bibinfo{title}{Which training methods for gans do actually
  converge?}, in: \bibinfo{booktitle}{ICML}.
\bibitem[{Ren et~al.(2021)Ren, Dey, Fishbaugh and Gerig}]{ren2021segmentation}
\bibinfo{author}{Ren, M.}, \bibinfo{author}{Dey, N.},
  \bibinfo{author}{Fishbaugh, J.}, \bibinfo{author}{Gerig, G.},
  \bibinfo{year}{2021}.
\newblock \bibinfo{title}{Segmentation-renormalized deep feature modulation for
  unpaired image harmonization}.
\newblock \bibinfo{journal}{IEEE Transactions on Medical Imaging}
  \bibinfo{volume}{40}, \bibinfo{pages}{1519--1530}.
\bibitem[{Ronneberger et~al.(2015)Ronneberger, Fischer and
  Brox}]{Ronneberger2015UNetCN}
\bibinfo{author}{Ronneberger, O.}, \bibinfo{author}{Fischer, P.},
  \bibinfo{author}{Brox, T.}, \bibinfo{year}{2015}.
\newblock \bibinfo{title}{U-net: Convolutional networks for biomedical image
  segmentation}, in: \bibinfo{booktitle}{MICCAI}.
\bibitem[{Sharma and Hamarneh(2020)}]{Sharma2020MissingMP}
\bibinfo{author}{Sharma, A.}, \bibinfo{author}{Hamarneh, G.},
  \bibinfo{year}{2020}.
\newblock \bibinfo{title}{Missing mri pulse sequence synthesis using
  multi-modal generative adversarial network}.
\newblock \bibinfo{journal}{IEEE Transactions on Medical Imaging}
  \bibinfo{volume}{39}, \bibinfo{pages}{1170--1183}.
\bibitem[{Shen et~al.(2021)Shen, Zhu, Wang, Xing, Pauly, Turkbey, Harmon,
  Sanford, Mehralivand, Choyke, Wood and Xu}]{Shen2021MultiDomainIC}
\bibinfo{author}{Shen, L.}, \bibinfo{author}{Zhu, W.}, \bibinfo{author}{Wang,
  X.}, \bibinfo{author}{Xing, L.}, \bibinfo{author}{Pauly, J.M.},
  \bibinfo{author}{Turkbey, B.}, \bibinfo{author}{Harmon, S.A.},
  \bibinfo{author}{Sanford, T.}, \bibinfo{author}{Mehralivand, S.},
  \bibinfo{author}{Choyke, P.L.}, \bibinfo{author}{Wood, B.J.},
  \bibinfo{author}{Xu, D.}, \bibinfo{year}{2021}.
\newblock \bibinfo{title}{Multi-domain image completion for random missing
  input data}.
\newblock \bibinfo{journal}{IEEE Transactions on Medical Imaging}
  \bibinfo{volume}{40}, \bibinfo{pages}{1113--1122}.
\bibitem[{Siegel et~al.(2019)Siegel, Miller and Jemal}]{Siegel2019CancerS2}
\bibinfo{author}{Siegel, R.L.}, \bibinfo{author}{Miller, K.D.},
  \bibinfo{author}{Jemal, A.}, \bibinfo{year}{2019}.
\newblock \bibinfo{title}{Cancer statistics, 2019}.
\newblock \bibinfo{journal}{CA: A Cancer Journal for Clinicians}
  \bibinfo{volume}{69}.
\bibitem[{Song and Ye(2021)}]{DBLP:journals/corr/abs-2106-09246}
\bibinfo{author}{Song, J.}, \bibinfo{author}{Ye, J.C.}, \bibinfo{year}{2021}.
\newblock \bibinfo{title}{Federated cyclegan for privacy-preserving
  image-to-image translation}.
\newblock \bibinfo{journal}{CoRR} \bibinfo{volume}{abs/2106.09246}.
\bibitem[{Wang et~al.(2020)Wang, Yurochkin, Sun, Papailiopoulos and
  Khazaeni}]{DBLP:conf/iclr/WangYSPK20}
\bibinfo{author}{Wang, H.}, \bibinfo{author}{Yurochkin, M.},
  \bibinfo{author}{Sun, Y.}, \bibinfo{author}{Papailiopoulos, D.S.},
  \bibinfo{author}{Khazaeni, Y.}, \bibinfo{year}{2020}.
\newblock \bibinfo{title}{Federated learning with matched averaging}, in:
  \bibinfo{booktitle}{{ICLR}}, \bibinfo{publisher}{OpenReview.net}.
\bibitem[{Wang et~al.(2018)Wang, Zhou, Wang, Yu, Zu, Lalush, Lin, Wu, Zhou and
  Shen}]{Wang2018LocalityAM}
\bibinfo{author}{Wang, Y.}, \bibinfo{author}{Zhou, L.}, \bibinfo{author}{Wang,
  L.}, \bibinfo{author}{Yu, B.}, \bibinfo{author}{Zu, C.},
  \bibinfo{author}{Lalush, D.S.}, \bibinfo{author}{Lin, W.},
  \bibinfo{author}{Wu, X.}, \bibinfo{author}{Zhou, J.}, \bibinfo{author}{Shen,
  D.}, \bibinfo{year}{2018}.
\newblock \bibinfo{title}{Locality adaptive multi-modality gans for
  high-quality pet image synthesis}.
\newblock \bibinfo{journal}{Medical image computing and computer-assisted
  intervention : MICCAI ... International Conference on Medical Image Computing
  and Computer-Assisted Intervention} \bibinfo{volume}{11070},
  \bibinfo{pages}{329--337}.
\bibitem[{Wu et~al.(2023)Wu, Wang, Zheng, Li, Alsaadi and Zeng}]{Wu2023AGGNAG}
\bibinfo{author}{Wu, P.}, \bibinfo{author}{Wang, Z.}, \bibinfo{author}{Zheng,
  B.}, \bibinfo{author}{Li, H.}, \bibinfo{author}{Alsaadi, F.E.},
  \bibinfo{author}{Zeng, N.}, \bibinfo{year}{2023}.
\newblock \bibinfo{title}{Aggn: Attention-based glioma grading network with
  multi-scale feature extraction and multi-modal information fusion}.
\newblock \bibinfo{journal}{Computers in biology and medicine}
  \bibinfo{volume}{152}, \bibinfo{pages}{106457}.
\bibitem[{Yang et~al.(2021)Yang, Sun, Yang and Xu}]{Yang2021AUH}
\bibinfo{author}{Yang, H.}, \bibinfo{author}{Sun, J.}, \bibinfo{author}{Yang,
  L.}, \bibinfo{author}{Xu, Z.}, \bibinfo{year}{2021}.
\newblock \bibinfo{title}{A unified hyper-gan model for unpaired multi-contrast
  mr image translation}, in: \bibinfo{booktitle}{International Conference on
  Medical Image Computing and Computer-Assisted Intervention}.
\bibitem[{Yu et~al.(2018)Yu, Zhou, Wang, Fripp and Bourgeat}]{Yu20183DCB}
\bibinfo{author}{Yu, B.}, \bibinfo{author}{Zhou, L.}, \bibinfo{author}{Wang,
  L.}, \bibinfo{author}{Fripp, J.}, \bibinfo{author}{Bourgeat, P.T.},
  \bibinfo{year}{2018}.
\newblock \bibinfo{title}{3d cgan based cross-modality mr image synthesis for
  brain tumor segmentation}.
\newblock \bibinfo{journal}{2018 IEEE 15th International Symposium on
  Biomedical Imaging (ISBI 2018)} , \bibinfo{pages}{626--630}.
\bibitem[{Yu et~al.(2020)Yu, Zhou, Wang, Shi, Fripp and
  Bourgeat}]{Yu2020SampleAdaptiveGL}
\bibinfo{author}{Yu, B.}, \bibinfo{author}{Zhou, L.}, \bibinfo{author}{Wang,
  L.}, \bibinfo{author}{Shi, Y.}, \bibinfo{author}{Fripp, J.},
  \bibinfo{author}{Bourgeat, P.T.}, \bibinfo{year}{2020}.
\newblock \bibinfo{title}{Sample-adaptive gans: Linking global and local
  mappings for cross-modality mr image synthesis}.
\newblock \bibinfo{journal}{IEEE Transactions on Medical Imaging}
  \bibinfo{volume}{39}, \bibinfo{pages}{2339--2350}.
\bibitem[{Yurochkin et~al.(2019)Yurochkin, Agarwal, Ghosh, Greenewald, Hoang
  and Khazaeni}]{DBLP:conf/icml/YurochkinAGGHK19}
\bibinfo{author}{Yurochkin, M.}, \bibinfo{author}{Agarwal, M.},
  \bibinfo{author}{Ghosh, S.}, \bibinfo{author}{Greenewald, K.H.},
  \bibinfo{author}{Hoang, T.N.}, \bibinfo{author}{Khazaeni, Y.},
  \bibinfo{year}{2019}.
\newblock \bibinfo{title}{Bayesian nonparametric federated learning of neural
  networks}, in: \bibinfo{booktitle}{{ICML}}, \bibinfo{publisher}{{PMLR}}. pp.
  \bibinfo{pages}{7252--7261}.
\bibitem[{Zhou et~al.(2021)Zhou, Liu and Duncan}]{Zhou2021AnatomyConstrainedCL}
\bibinfo{author}{Zhou, B.}, \bibinfo{author}{Liu, C.}, \bibinfo{author}{Duncan,
  J.S.}, \bibinfo{year}{2021}.
\newblock \bibinfo{title}{Anatomy-constrained contrastive learning for
  synthetic segmentation without ground-truth}, in:
  \bibinfo{booktitle}{MICCAI}.
\bibitem[{Zhu et~al.(2017)Zhu, Park, Isola and Efros}]{zhu2017unpaired}
\bibinfo{author}{Zhu, J.Y.}, \bibinfo{author}{Park, T.},
  \bibinfo{author}{Isola, P.}, \bibinfo{author}{Efros, A.A.},
  \bibinfo{year}{2017}.
\newblock \bibinfo{title}{Unpaired image-to-image translation using
  cycle-consistent adversarial networks}, in: \bibinfo{booktitle}{Proceedings
  of the IEEE international conference on computer vision}, pp.
  \bibinfo{pages}{2223--2232}.
\bibitem[{Zuo et~al.(2021)Zuo, Zhang and Yang}]{Zuo2021DMCFusionDM}
\bibinfo{author}{Zuo, Q.}, \bibinfo{author}{Zhang, J.}, \bibinfo{author}{Yang,
  Y.}, \bibinfo{year}{2021}.
\newblock \bibinfo{title}{Dmc-fusion: Deep multi-cascade fusion with
  classifier-based feature synthesis for medical multi-modal images}.
\newblock \bibinfo{journal}{IEEE Journal of Biomedical and Health Informatics}
  \bibinfo{volume}{25}, \bibinfo{pages}{3438--3449}.

\end{thebibliography}


\bio{authors/jinbao-wang1}
Jinbao Wang received the Ph.D. degree from the University of Chinese Academy of Sciences (UCAS) in 2019. He is currently a Research Assistant Professor with the Southern University of Science and Technology (SUSTech), Shenzhen, China. His research interests include machine learning, computer vision, image anomaly detection, and graph representation learning.
\endbio

\vskip20pt

\bio{authors/xie1}
Guoyang Xie received the Bachelor and MPhil Degrees from University of Electronic Science and Technology of China, Hong Kong University of Science and Technology in 2009 and 2013, respectively. He is pursuing the PhD degree from University of Surrey. Prior to that, he was the Principle Perception Algorithm Engineer in Baidu and GAC, respectively. His research interests include anomaly detection, medical imaging, neural architecture search and federated learning.
\endbio

\bio{authors/huang}
Yawen Huang received the M.Sc. and Ph.D. degrees from the Department of Electronic and Electrical Engineering, The University of Sheffield, Sheffield, U.K., in 2015 and 2018, respectively. She is currently a Senior Scientist of Tencent Jarvis Laboratory, Shenzhen, China. Her research interests include computer vision, machine learning, medical imaging, deep learning, and practical AI for computer aided diagnosis.
\endbio

\bio{authors/jiayi-lyu1}
Lyu Jiayi, born in 1999, graduated in 2021 from Capital Normal University with a bachelor's degree in computer science and technology. She is now pursuing a Ph.D. in computer applications at the School of Engineering Science, Chinese Academy of Sciences.\\
\endbio

\bio{authors/feng-zheng}
Feng Zheng (Member, IEEE) received the Ph.D. degree from The University of Sheffield, Sheffield, U.K., in 2017. He is currently an Assistant Professor with the Department of Computer Science and Engineering, Southern University of Science and Technology, Shenzhen, China. His research interests include machine learning, computer vision, and human-computer interaction.
\endbio
\bio{authors/yefeng-zheng}
Yefeng Zheng (Fellow, IEEE) received the B.E. and M.E. degrees from Tsinghua University, Beijing, in 1998 and 2001, respectively, and the Ph.D. degree from the University of Maryland, College Park, MD, USA, in 2005. After graduation, he joined Siemens Corporate Research, Princeton, NJ, USA. He is currently the Director and the Distinguished Scientist of Tencent Jarvis Laboratory, Shenzhen, China, leading the company’s initiative on Medical AI. His research interests include medical image analysis, graph data mining, and deep learning. Dr. Zheng is a fellow of the American Institute for Medical and Biological Engineering (AIMBE).
\endbio
\bio{authors/yaochu-jin}
Yaochu Jin received the B.Sc., M.Sc., and Ph.D. degrees from Zhejiang University, Hangzhou, China, in 1988, 1991, and 1996, respectively, and the Dr.-Ing. degree from Ruhr University Bochum, Germany, in 2001. 

He is presently an Alexander von Humboldt Professor for Artificial Intelligence endowed by the German Federal Ministry of Education and Research, Chair of Nature Inspired Computing and Engineering, Faculty of Technology, Bielefeld University, Germany. He is also a Distinguished Chair, Professor in Computational Intelligence, Department of Computer Science, University of Surrey, Guildford, U.K. He was a “Finland Distinguished Professor” of University of Jyväskylä awarded by the Academy of Science and Finnish Funding Agency for Innovation, Finland, and “Changjiang Distinguished Visiting Professor” of Northeastern University, awarded by the Ministry of Education, China. His main research interests include human-centered learning and optimization, synergies between evolution and learning, and evolutionary developmental artificial intelligence. 

Prof. Jin is the President-Elect of the IEEE Computational Intelligence Society and the Editor-in-Chief of Complex \& Intelligent Systems. He was named by Clarivate as a “Highly Cited Researcher” from 2019 to 2022 consecutively. He is a Member of Academia Europaea and Fellow of IEEE.

\endbio



\end{document}